# Evolution of Activation Functions: An Empirical Investigation

ANDREW NADER, Department of Computer Science and Mathematics, Lebanese American University
DANIELLE AZAR(CORRESPONDING AUTHOR), Department of Computer Science and Mathematics, Lebanese American University

The hyper-parameters of a neural network are traditionally designed through a time consuming process of trial and error that requires substantial expert knowledge. Neural Architecture Search (NAS) algorithms aim to take the human out of the loop by automatically finding a good set of hyper-parameters for the problem at hand. These algorithms have mostly focused on hyper-parameters such as the architectural configurations of the hidden layers and the connectivity of the hidden neurons, but there has been relatively little work on automating the search for completely new activation functions, which are one of the most crucial hyperparameters to choose. There are some widely used activation functions nowadays which are simple and work well, but nonetheless, there has been some interest in finding better activation functions. The work in the literature has mostly focused on designing new activation functions by hand, or choosing from a set of predefined functions while this work presents an evolutionary algorithm to automate the search for completely new activation functions. We compare these new evolved activation functions to other existing and commonly used activation functions. The results are favorable and are obtained from averaging the performance of the activation functions found over 30 runs, with experiments being conducted on 10 different datasets and architectures to ensure the statistical robustness of the study.

## 1 INTRODUCTION

The output of a neuron in the standard model of a neural network can be represented as the composition of a linear function and an arbitrary scalar function, where the linear function is the euclidean product of the inbound weights and inputs to the neuron, and the arbitrary function is a hyper-parameter to the neural network called the activation function. Since a neural network may itself be represented as a composition of functions, a neural network with a linear activation function in all neurons will only be able to learn linear relationships. Thus, in general, we require the activation function to be non-linear. In fact, the Universal Approximation Theorem [Csáji

Authors' addresses: Andrew Nader, Department of Computer Science and Mathematics, Lebanese American University, Byblos, Lebanon, andrew.nader@lau.edu; Danielle Azar(Corresponding author), Department of Computer Science and Mathematics, Lebanese American University, Byblos, Lebanon, danielle.azar@lau.edu.lb.

2001] states that a neural network with one hidden layer and a bounded, non-constant, continuous (and thus, non-linear) activation function is a universal function approximator. The rest of the mathematical properties that make for a good activation function are not yet completely known, and they may, in fact, be problem dependent, but there are some properties such as monotonicity and continuous differentiability which have been shown to work well in practice. The most commonly used activation function is the rectified linear unit (ReLU) $f(x) = max(0, x)$ [Nair and Hinton 2010],which has been shown to enable the training of deep neural networks [Krizhevsky et al. 2012]. Even though ReLU is simple and works well, there has been some research in designing new activation functions, but the research has mostly focused on designing new activation functions by hand by following some commonly used heuristics.

Inspired by the recent success of Neural Architecture Search (NAS) methods [Elsken et al. 2019; Hutter et al. 2019; Zoph et al. 2018], which aim to automate the design of a neural network, we propose an evolutionary algorithm to automatically discover new activation functions that perform well. While NAS methods can come up with completely new architectures, most implementations tend to sample from a small, predefined set of hand-designed activation functions [Feurer et al. 2015; Mendoza et al. 2018; Sun et al. 2018]. We aim to show that it is worth augmenting NAS methods by adding the search for new activation functions and we note that it is relatively simple to add our algorithm to existing evolutionary NAS methods. We keep the architecture fixed over each run of our algorithm, to reduce the number of confounding variables and properly study the effect of the search. To ensure the statistical robustness of our findings, we test the algorithm on 10 different datasets and architectures. Our experiments consist of 3 classification problems on image-based datasets, 4 classification problems on non-image-based datasets, and 3 regression problems. At the end of the algorithm, we report the mean and standard deviation of the performance metrics for the best three activation functions in the final generation over 30 runs, and the mean and standard deviation of the performance metrics of the ReLU, ELU, and SELU over 30 runs, using the same architecture. In addition, we are interested in studying the shape of the best activation functions on each dataset. To do this, we plot the evolution of 4 different properties, which are the Monotonicity, Zero on Non-Negative Real Axis, Upper Unbounded, and Lower Unbounded properties. These properties describe the shape of the activation functions being processed by the algorithm throughout its run to see what type of activation function is being favored, and the reasoning behind each property is presented in Section 3.3. The rest of this paper is organized as follows: we review the literature on activation functions in Section 2, we describe the experimental methodology in Section 3, we present the results and analysis in Section 4, and finally we conclude and identify multiple lines of avenue for future work in Section 5.

## 2   RELATED WORK

Before discussing more recently found activation functions, we first describe some of the more traditionally used activations. The activation functions used before the discovery of ReLU were mostly sigmoidal in shape (that is, "S" shaped) : some of the commonly used sigmoidal functions were the hyperbolic tangent defined by $f(x) = \frac{e^x - e^{-x}}{e^x + e^{-x}}$, and the logistic function (which is commonly simply called the sigmoid) defined by $f(x) = \frac{1}{1+e^{-x}}$. These activation functions are inspired by neurons in the brain, which may be modelled by the Heaviside step function (either the neuron fires, or not). The sigmoidal functions can provide a differentiable approximation to binary step functions. The main problem with these functions is that their derivative tends to zero as $\mid x \mid \rightarrow \infty$, which hurts the speed of learning of the gradient-based optimization of neural networks. This problem is exacerbated in deep networks: as the gradient propagates and multiplies throughout the layers, the update to the parameters of the network by the gradient becomes negligible. This results in a deep network that barely learns from the training examples. The rectified linear unit (ReLU) defined by $f(x) = max(0, x)$ solves this problem by having a gradient of one if the input is positive, and zero otherwise. ReLU is also fast to compute, since the function does not require the computation of an exponential or any other computationally expensive operation. ReLU was first shown to provide major improvements in [Glorot et al. 2011]. The authors argue that ReLU is more biologically plausible than sigmoidal activations since it introduces sparsity in the neural network because of its hard zeroes when one moves to the negative part of the x-axis: A prevalent theory is that the brain learns distributed sparse representations, in the sense that for each task presented to it, only a small percentage of the neurons fire [Attwell and Laughlin 2001]. The authors also argue that sparse representations may lead to multiple advantages, such as the higher likelihood of sparse representations being linearly separable, and that sparse representations do not suffer from the highly entangled information in dense representations: In dense representations, any change to the input vector modifies most of the entries in the representation vector, which may lead to "tangled" information representation. The authors then experimentally validate their ideas, and show that ReLU outperforms the tanh and the softplus (which is a differentiable approximation to ReLU) on 4 datasets. The softplus is almost identical to ReLU, but it is differentiable around the origin. Thus, it is almost zero around the origin, but it does not have the abrupt jump that ReLU has. The authors hypothesize that this means that softplus does not induce the sparsity that ReLU does, and this may explain the difference in performance.However, it is unclear whether sparse representations are actually useful to the process of learning in neural networks [Xu et al. 2015]. In addition, there has been some debate as to whether the sparsity is actually required in the brain itself: sparse representations lead to worse generalization errors, it could be that the experiments that hypothesize that the brain learns sparse representations are flawed, and there has also been some experimental evidence that contradicts the claim that sparse coding is a governing principle in the brain [Spanne and Jörntell 2015]. Another seminal result related to ReLU in the research on activation functions can be found in [Nair and Hinton 2010]. The authors show that the performance of a Restricted Boltzmann Machine can be improved by replacing the binary hidden units by NReLU, which is obtained by adding Gaussian Noise to the positive part of ReLU. The authors explain that the increase in performance is due to the fact that NReLU preserves the information about the relative intensities. One potential problem with ReLU is the dying neurons problem [Lu et al. 2019]: the zero part can sometimes cause some neurons to "die", in the sense that they never activate again for any input. One way this could happen is if the neural network learns a large negative bias for a neuron, which pushes the input to be less than zero. One must distinguish between the dying neurons problem and the sparse representation offered by ReLU: for a sparse representation, only a small subset of the neurons activate on a particular input, but that does not mean that there are some neurons that *never* activate for any input: the small subset that activates is different for each input. Thus, each neuron is contributing to the network, depending on the input used. The dying ReLU problem is different in that there are neurons that do no activate no matter what the input is. This means that they are contributing nothing to the Neural Network.

One way to solve the dying neuron problem which occurs with ReLU is by to introduce a linear part with a small slope instead of the zero. This is done by the Leaky ReLU, first introduced in [Maas et al. 2013]. The authors show that ReLU offers an advantage over the sigmoid for Neural Network Acoustic models, and they also introduce the Leaky ReLU which is defined by $f(x) = max(0.01x, x)$. Leaky ReLU is identical to ReLU when $x > 0$, and it replaces the zero by $0.01x$ when $x < 0$. This sacrifices sparsity for the escape from the dying ReLU problem. The Leaky ReLU is shown to perform similarly to the ReLU, which leads to the conclusion that it may offer the potential to escape the dying ReLU problem without sacrificing performance. It is worth noting that the authors in [Xu et al. 2015] try to increase the slope of the negative part of Leaky ReLU to 1/5.5 (Leaky ReLU becomes $f(x) = max(\frac{x}{5.5}, x)$) and gain an improvement over the standard Leaky ReLU and ReLU, which perform similarly.

A commonly used variant on ReLU is the Exponential Linear Unit (ELU), introduced in [Clevert et al. 2015]. The authors explain that activation functions like Leaky ReLU and the PReLU [He et al. 2015] are better choices than ReLU because their mean activation is closer to zero because of the presence of negative values (For a discussion of the effects of zero centering, see [Le Cun et al. 1991; LeCun et al. 2012; Raiko et al. 2012; Schraudolph 1998]), but that the potential for large negative values may decrease their robustness to noise when they are deactivated (when they output negative values). ELU provides negative values to push the mean closer to zero, and its negative values saturate, so it provides the best of both worlds. Of course, the disadvantage is that it requires the computation of an exponential. The authors then run experiments, and show that ELU performs very well, achieving a state of the art result on the CIFAR-100 dataset. The authors in [Klambauer et al. 2017] introduce a way to automatically push the means of the activations towards 0 and the variance towards the unit variance by using the Self Normalizing Exponential Linear Unit (SELU) as the activation function in a neural network. SELU neural networks do not suffer from the exploding or vanishing gradient problem, and this property shows its utility in the experiments, where the authors find that SELU outperforms many other activations on extremely deep architectures, including some that have had explicit normalization using techniques such as Batch Normalization. The authors in [Hendrycks and Gimpel 2016] introduce a new function called the Gaussian Error Linear Unit (GELU). GELU is defined by $f(x) = xP(X \le x)$. where $X \sim N(\mu, \sigma^2)$. The authors define a stochastic regularizer that takes into account the input and randomly selects the identity transformation or the zero transformation, and they show that GELU is the expected transformation of this regularizer. $\mu$ and $\sigma^2$ can be tuned, but the authors simply set them to zero and one for their experiments. They find that GELU outperforms ReLU and ELU over multiple datasets. At the end of the paper, they note that GELU is closely related to ReLU, in that GELU $\rightarrow$ ReLU as $\sigma \rightarrow 0$ if $\mu = 0$, but the activations are different enough in the general case because GELU is non-convex and non-monotonic, which they hypothesize might help it learn more difficult functions. Some experiments on non-monotonic activation functions can also be found in [Sopena et al. 1999].

Researchers at Google Brain attempt to automatically search for activation functions using a Recurrent Neural Network controller [Ramachandran et al. 2017]. They define an activation function as a collection of "components" put together, where the components are chosen by a recurrent neural network controller which is trained using a reinforcement learning algorithm. They found multiple activation functions that performed well, but the activation function that they decided to go with as it seemed to perform best is $f(x) = x$ sigmoid($\beta x$), where $\beta$ can be chosen arbitrarily. The function is dubbed Swish. They then proceed to test Swish on different architectures and different datasets such as CIFAR-10, CIFAR-100, and Imagnet, and they compare the median performance across five different runs to other activations. The activations that they compare Swish to are ReLU, LReLU, PReLU [He et al. 2015], Softplus, ELU, SELU, and GELU. The experiments show that Swish consistently outperforms ReLU, and that Swish outperforms or matches most of the rest of the activation functions on all models. On the evolutionary computation side of things, two very recent papers came out at the GECCO 2020 conference [Bingham et al. 2020; Nader and Azar 2020], both of which showed that evolutionary algorithms succeed in evolving good activation functions. We view the current paper as adding to the empirical evidence provided by these two papers that it is worth searching for completely new activation functions. Finally, while the automated search proposed in [Bingham et al. 2020; Nader and Azar 2020; Ramachandran et al. 2017] and the current work uses the same activation function for all the neurons in the hidden layers, the work in [Cui et al. 2019; Shabash and Wiese 2018] evolves heterogeneous neural networks where each neuron may have a different activation function; these functions are chosen from a predefined set and are not created by the evolutionary algorithm itself. In addition, the authors adjust the weights of the neural network through the evolutionary algorithm, so the activation functions do not have to be differentiable. The authors find that this approach performs well, with the heterogeneous neural network outperforming standard homogeneous ANNs. In this work, we decided to use the same activation function in all hidden neurons in order to be able to analyze the shape of the evolved function; however, in future work, it would be interesting to extend our approach by designing an algorithm that can both create heterogeneous networks and search for completely new activation functions.

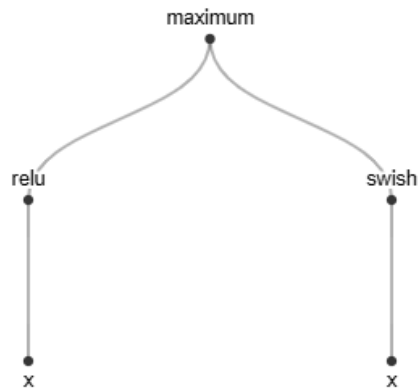

Fig. 1. Activation Tree Example.

## 3 METHODOLOGY

Tree-based structures offer a natural representation for mathematical functions, and thus genetic programming is a suitable evolutionary technique for evolving activation functions. An activation function may also be heavily dependent on the probability distribution associated with the initial random weights of the neural networks, so we co-evolve a weight initialization gene with the activation function. Thus, our chromosome is of the form $< f, w >$ where $f$ is the activation function tree and $w$ is a nominal variable representing the weight initialization scheme used. We first describe the activation function tree representation and weight initialization scheme in greater detail, and we move on to a discussion of the experimental parameters used.

### 3.1 Chromosome Encoding

The first gene $f$ represents the activation function which is represented as a tree using a genetic programming approach. We show an example of such a tree in Figure 1 where the leaf nodes represent the input to the activation function, and the internal nodes consist of either unary or binary operations on their children nodes. Thus, the tree shown in Figure 1 encodes the activation function max(ReLU(x),Swish(x)). It is important to see that two activation functions, represented as trees, can exchange useful information through crossover, and hence the resulting search algorithm is not a random one. For this, and for illustration purposes, we consider the activation function trees composed of the functions $f_1(x)$, $g_1(x)$ and ReLU(x) pictured in Figure 2.

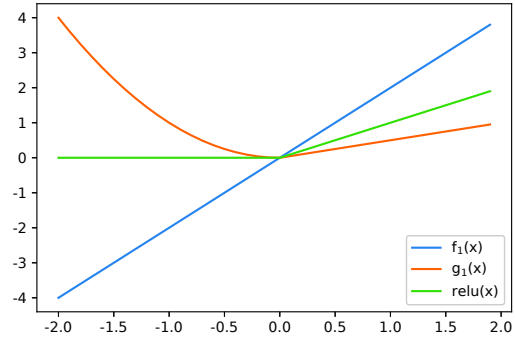

Fig. 2. Plot of $f_1$, $g_1$ and ReLU.

Consider the activation function trees shown in Figures 3a and 3b, which incorporate the functions of Figure 2 as internal nodes. The activation function tree $f(x)$ defines a function that is equal to zero when x is non-positive, and equal to $f_1(x)$ when x is positive, while the activation

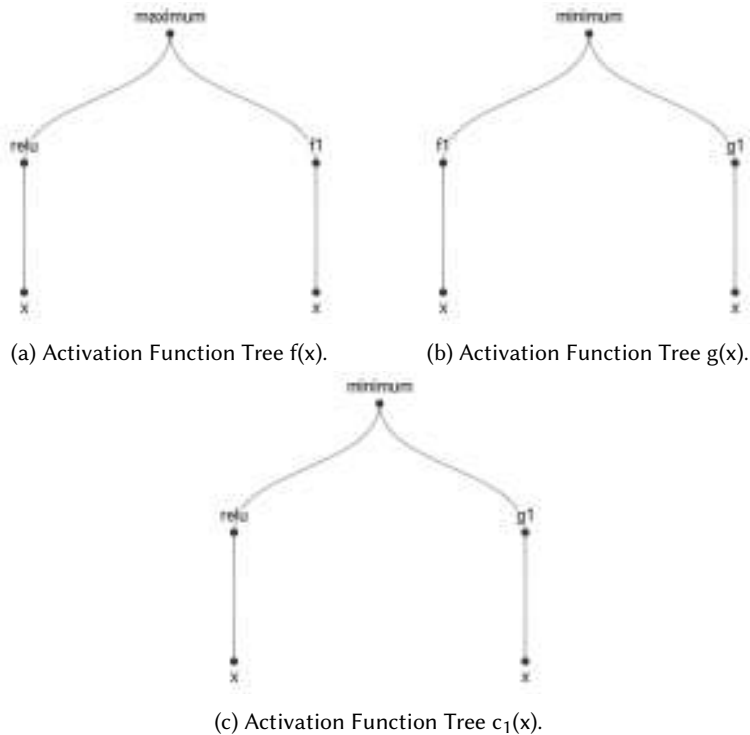

(a) Activation Function Tree f(x).          (b) Activation Function Tree g(x).

(c) Activation Function Tree $c_1(x)$.

Fig. 3. Activation Function Trees f(x), g(x) and $c_1(x)$. $c_1(x)$ is one of the trees resulting from crossover between f(x) and g(x).

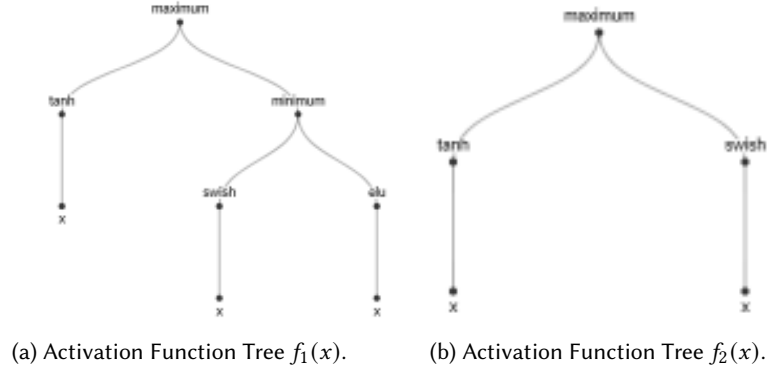

(a) Activation Function Tree $f_1(x)$.                    (b) Activation Function Tree $f_2(x)$.

Fig. 4. Illustration of the shrink mutation. The activation function tree $f_1(x)$ is mutated by randomly choosing a branch node (minimum, in this case), and replacing it with one of its arguments (swish(x), in this case). $f_2(x)$ is the resulting tree.

function tree $g(x)$ defines a function that is equal to $f_1(x)$ when x is non-positive, and equal to $g_1(x)$ when x is positive. Let us assume that both $g(x)$ and $f(x)$ perform very well, that $g(x)$ performs better than $f(x)$, and that hard zeroes on $\mathbb{R}^-$ are a very useful property for *all* activation functions, which partly explains the good performance of $f(x)$. Then, by performing crossover on $f(x)$ and $g(x)$ and exchanging the left sub-trees rooted at the first level, one of the activation function trees obtained will be $c_1(x)$, shown in Figure 3c. Since we know that hard zeroes are useful for all activation functions, then this function will necessarily be better than $g(x)$. It is trivial to come up with similar examples that show how other activation function properties such as non-monotonicity can also be transferred through crossover. Thus, it is obvious that useful properties in activation functions can be transferred through genetic programming. As discussed before, the internal nodes in the trees can be either unary or binary operations on their children nodes. We use the following tensorflow [Abadi et al. 2015] operations in our evolutionary algorithm:

Unary operations: *relu, elu, sigmoid, tanh, swish, sin, cosin, atan ,sinh, cosh, leaky relu, softplus, erf,* and *absolute value.*

Binary operations: *add, subtract, multiply, maximum,* and *minimum.*

The second gene $w$ in our chromosome is a nominal variable that designates the weight initialization scheme of a neural network. We use the following weight initialization schemes in our evolutionary algorithm:

Weight initialization schemes: *random normal, random uniform, truncated normal, variance scaling, orthogonal* [Saxe et al. 2013], *lecun uniform* [LeCun et al. 2012], *lecun normal* [Klambauer et al. 2017; LeCun et al. 2012], *glorot uniform* [Glorot and Bengio 2010], *glorot normal*[Glorot and Bengio 2010], *he normal* [He et al. 2015], and *he uniform* [He et al. 2015]. We note that we use a different seed for the random number generator every time it is invoked.

## 3.2 Experimental Setup

We keep a population size of 100 individuals for 50 generations and we use an elitism strategy of keeping the best 4 individuals in all of our experiments. We use rank selection, a crossover rate of 80% and a mutation rate of 5% for our evolutionary algorithm. The crossover operator is simple: given two parent chromosomes $p_1 = <f_1, w_1>$ and $p_2 = <f_2, w_2>$, we first perform a single point chromosome crossover to obtain the two children chromosomes $c_1 = <f_1, w_2>$ and $c_2 = <f_2, w_1>$. The two activation function trees then undergo crossover with the *leaf-biased* one-point crossover operator implemented in the Python DEAP library [Fortin et al. 2012]. We choose this genetic programming operator since this parameter does nfot affect the performance significantly. As a matter of fact, in [Nader and Azar 2020], it is shown that on different data sets, a different variational operator was found to perform best and hence, no single one showed to be better than all others. So, we concluded that they are all equally valid. The leaf-biased crossover operator is identical to the standard one-point sub-tree exchange used in genetic programming (illustrated in Figure 3), except for the fact that it only has a 10% chance of choosing a leaf node as the crossover point.

Since exchanging two leaf nodes between trees is equivalent to refraining from doing any crossover, we opt for the leaf-biased operator. We use a *shrink* mutation operator, also implemented in DEAP which randomly chooses a branch and replaces it with a random element of one of the branch arguments. This mutation operator is illustrated in Figure 4. We choose to constrain the initial depth of the trees between 1 and 4, and we use a static bloat control where a child tree is replaced by a randomly chosen parent if it exceeds a depth of 10, to bias our search towards activation functions that are quick to optimize.

Once an activation function is found, it is applied to every neuron in the hidden layers of the network and the latter is designed and trained using the Keras framework [Chollet et al. 2015] and ADAM optimizer [Kingma and Ba 2014] with a default batch size of 32. We also use a balanced class weight in all of our experiments. If we were to evaluate the fitness on the training set, we would bias our search towards individuals that over-fit the training set easily. Furthermore, in order to avoid evaluating the fitness on the unseen testing data, we keep a separate validation set to evaluate the fitness of the individuals found by the evolutionary algorithm. For the classification experiments, we use the $F_1$ measure as a fitness metric if the target in the dataset is binary, and we use the categorical accuracy as a metric if the target consists of more than two classes. For the case of regression, we use the mean squared error as a fitness metric. We keep 10% of the training set as a validation set for all of our datasets.

We switch between two different techniques depending on whether the evolutionary search for individuals has terminated or not:

(1) The evolutionary search has not yet terminated: we use early stopping with a tolerance of 10 epochs on the validation set (we stop the training and restore the best weights if the performance does not increase for 10 epochs) and the individuals are trained for a maximum of 50 epochs. The metric used for early stopping is the same as the fitness metric chosen for the dataset that we are dealing with. In this case, the split for the validation set is random to prevent the evolutionary algorithm from finding functions that implicitly over-fit on a particular validation set, where we use "implicit over-fitting" in the sense that always choosing the same validation set can cause the evolutionary algorithm to favor activation functions that only perform well on this particular set, even though they have not seen its data during the neural network training phase.

(2) The evolutionary search has terminated, i.e we have found the final population which we are about to train and test on the final unseen testing data: we use model check-pointing on the validation set (we keep track of the best weights without any early stopping) and the individuals are trained for 100 epochs. As in the case of early stopping, the metric for the model check-pointing is the same as the fitness metric chosen for the individuals. In this case, the split for the validation set is deterministic: since the evolutionary search has terminated, we do not have to worry about it implicitly over-fitting on a particular validation set. We choose a deterministic split to ensure that we do not get inaccurate results because some individual got lucky and obtained a validation set that was of higher quality than the others. We test each individual in the final population 30 times and report the mean and standard deviation of its performance. We choose to keep the train/validation/test split identical across the 30 runs. Thus, when testing the final population, the randomness across different individuals and repeated runs of the same individual arises from algorithmic considerations such as random weight initialization and not from the particular split of a dataset.

The final population is obtained by choosing the 10 individuals with the highest fitness at the 100th epoch and getting the distinct individuals among these 10. For each data set, we choose from the final population the three best individuals evaluated on the validation set, we report their performance on the test set, and we compare them to three activation functions as a baseline: the ReLU with the Glorot uniform weight initialization scheme, ELU with the He normal weight initialization scheme, and SELU with the LeCun normal weight initialization scheme since these are commonly used combinations of activation functions/weight initialization schemes.

### 3.3 Evolution of Activation Function Properties

We study the evolution of activation function *properties* throughout the run of the evolutionary algorithm. The reason for doing this is to determine the shape of the activation functions that are performing well. For example, if 10% of the activation functions are monotonic at the start of the evolutionary algorithm, and this percentage steadily increases to 90% in the final generation, then this provides solid evidence that activation functions that are monotonic in shape are a good choice for this *particular* problem. Of course, it would not definitively imply that the *only* good activation functions are monotonic, but this evidence is still useful for analyzing what constitutes a good activation function and what constitutes a bad one. We study the following activation function properties:

(1) **Monotonicity**: this is a very commonly used heuristic for activation functions. The argument for monotonicity is that if the function is not monotonic, increasing the weight of a neuron through back-propagation might cause it to output lower values, thus decreasing the "importance" of the neuron, which is the opposite of the intended effect. In addition, the error surface of a single layer neural network associated with a monotonic activation function is convex [Wu 2009].

(2) **Zero on Non-Negative Real Axis**: this property checks if a function is zero when $x \leq 0$, like ReLU. We study this property to see if sparsity is useful in neural networks.

(3) **Upper Unbounded**: this property checks if a function $f(x) \to \infty$ as $x \to \infty$. Activation functions that are not upper unbounded like the logistic function have very small derivatives for large values of $x$, which hurts the speed of learning.

(4) **Lower Unbounded**: this property checks if a function $f(x) \to -\infty$ as $x \to -\infty$. As explained in [Clevert et al. 2015], activation functions such as the Leaky ReLU, perform better than ReLU because the presence of negative values helps push the mean closer to zero, but the potential for extremely large negative values decreases their robustness to noise. The lower unbounded property checks if the activation function has the potential for extremely large negative values.

## 4   RESULTS AND ANALYSIS

Our experiments consist of four experiments on classification problems that are not image-based, i.e they consist of multi-variate data, three experiments on multi-variate regression datasets, and three experiments on image-based classification problems. The architecture of the neural network is chosen randomly and is different for each dataset in order to see if our algorithm offers good results for a wide range of architectures. The datasets and architectures chosen are modest in size to keep the computational time required within acceptable bounds. We first discuss the results on the multi-variate classification datasets, we move on to the results on the regression datasets, and finally we present the results on the image-based classification datasets. In each case, we discuss the statistical significance of the results on every data set.

### 4.1   Results on Multi-variate Classification Datasets

The four datasets chosen for this experiment are the Electricity, Magic Telescope, Robot Navigation, and EEG Eye State datasets which are very popular datasets extracted using the Python package provided by the OpenML repository [Feurer et al. 2019; Vanschoren et al. 2013]. We perform a 75/25 train/test split for all four datasets. The datasets used are summarized in Table 1.

Table 1.  Description of the datasets used for the multi-variate classification experiments

| Dataset | № Instances | № Features | № classes | Minority/Majority Ratio (if binary target) |
|---------|-------------|------------|-----------|---------------------------------------------|
| Electricity | 45,312 | 9 | 2 | 73.733% |
| Magic Telescope | 19,020 | 12 | 2 | 54.23% |
| Robot Navigation | 5,456 | 25 | 4 | N/A |
| EEG Eye State | 14,980 | 15 | 2 | 81.42% |

The neural networks used for these datasets are standard feed-forward neural networks. Each neural network has a different combination of dropout [Srivastava et al. 2014] rates and/or $L_2$ regularization to prevent over-fitting. When we used dropout or $L_2$ regularization, we applied them to all of the layers except the first and last layer to prevent adding noise to the input and output of the network. We chose the dropout rates and whether to use $L_2$ regularization randomly in order to test multiple combinations. We encode categorical variables using a one-hot encoding scheme, and we also use a randomly chosen preprocessing method for the features in each dataset, where the preprocessing methods available are the MinMaxScaling to the [0,1] range and the StandardScaling to the [-1,1] range. For binary classification problems, we use a single output neuron with a sigmoid activation function and a binary cross-entropy loss, while for multi-class classification problems the number of output neurons is equal to the number of classes, and the activation function and loss are the softmax and categorical cross-entropy respectively. The architectures of the neural networks for each dataset and the method of preprocessing used is shown in Table 2.

Table 2.  Experimental Parameters for the Multi-variate Classification Datasets

| Dataset | № Hidden Layers | № Neurons per Layer | Dropout Rate | $L_2$ Regularization | Scaling Method |
|---------|-----------------|---------------------|--------------|----------------------|----------------|
| Electricity | 4 | 50 | 20% | Yes | StandardScaling |
| Magic Telescope | 2 | 100 | 50% | Yes | MinMaxScaling |
| Robot Navigation | 4 | 100 | 10% | No | StandardScaling |
| EEG Eye State | 3 | 30 | 0% | No | StandardScaling |

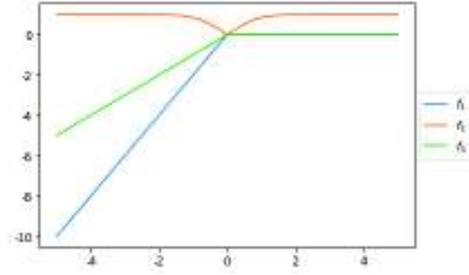

(a) Plot of the best 3 activation functions found on the Electricity dataset

- $f_1(x) = x - |x|$
- $f_2(x) = \text{erf}(|x|)$
- $f_3(x) = x - \text{relu}(x)$

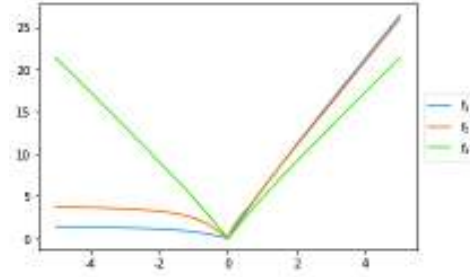

(b) Plot of the best 3 activation functions found on the Magic Telescope dataset

- $f_x1(x) = |\max(|x + \text{atan}(x)|, \text{atan}(\text{leaky-relu}(x) + |x|)) + \max(x, x + \max(2x + |x|, x + \text{atan}(\max(\text{atan}(x), |x|))))|$
- $f_2(x) = |\max(\text{atan}(2x + |x| + \text{relu}(x) + \text{leaky-relu}(x)), |x + \max(x, \text{atan}(x)|) + \max(x, \text{erf}(\max(|x|, \text{erf}(x))) + \max(2x + \max(x, \text{erf}(x)), x + \text{atan}(|x|)))|$
- $f_3(x) = |4x + \text{atan}(x)|$

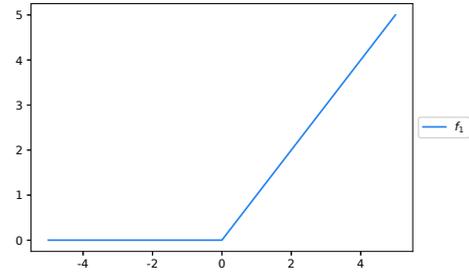

(c) Plot of the best activation function found on the Robot Navigation dataset

- $f_1(x) = \text{relu}(x)$

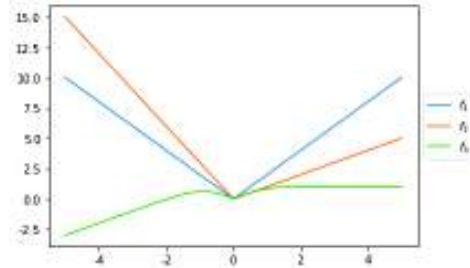

(d) Plot of the best 3 activation functions found on the EEG Eye State dataset

- $f_1(x) = 2|x|$
- $f_2(x) = 2|x| - x$
- $f_3(x) = \text{erf}(|x|) + x - \max(\tanh(x), x)$

Fig. 5. Plots of the best activation functions found for the multi-variate classification problems.

### 4.1.1 Best Three Functions Found for the Multi-Variate Classification Datasets:
The plots for the best 3 activation functions found for each of the datasets are shown in figures 5a, 5b, 5c, and 5d.

We can see that many of the activation functions in figures 5a and 5d are composed of approximately piece-wise linear functions. Even though these functions share this structural similarity, they also differ in some aspects such as the fact that $f_1$ and $f_3$ in Figure 5a are zero on a large subset of their domain (when $x \geq 0$), while $f_1$ and $f_2$ in Figure 5d are non-zero almost everywhere. It is interesting to note that the functions in Figure 5a are zero on $\mathbb{R}^+$ and non-zero on $\mathbb{R}^-$, which is the opposite of ReLU. The functions which are not piecewise-linear in figures 5a and 5d ($f_2$ and $f_3$, respectively) are highly surprising, first because of their unusual shape when compared to commonly used activation function, and second because they look somewhat similar, indicating that this is a potentially good activation function in general and not just for a very specific dataset. The functions found on the Magic Telescope dataset are shown in Figure 5b, and we can see that $f_1$ and $f_3$ have a linear shape when $x \geq 0$, while $f_2$ has a small bump in that domain although we suspect that this bump is insignificant. All of the functions are positive when $x < 0$, although $f_1$ and $f_2$ take on bounded values in that range while $f_3$ is unbounded. The algorithm converged to the ReLU function on the Robot Navigation dataset, as shown in Figure 5c, which is an indication that our algorithm is able to recover basic activation functions such as the ReLU if they perform well. This is a positive sign, as it means that our algorithm is a good candidate for automating the search for activation functions without being biased against well known activation functions as it is able to converge towards them if they perform better than more complicated functions, while it is also able to find new activation functions if they happen to outperform the base functions on a particular dataset. The means and standard deviations of the performance of the best activation functions found are shown in tables 3, 4, 5, 6 and plotted in Figure 6.

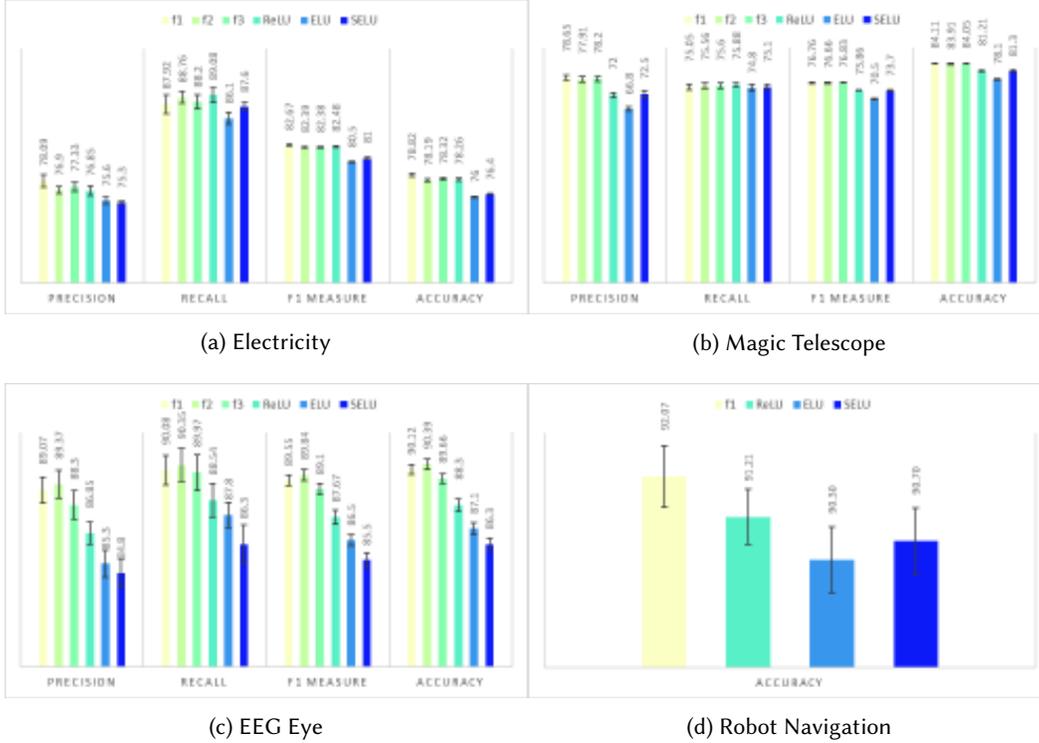

(a) Electricity                      (b) Magic Telescope

(c) EEG Eye                      (d) Robot Navigation

Fig. 6. Bar plots showing results (precision, recall, $F_1$ measure and accuracy) on the multi-variate classification datasets

Table 3 shows that $f_1$ and $f_3$ outperform the baseline functions on the accuracy front, and $f_1$ obtains a higher $F_1$ measure than all of the baseline functions for the electricity dataset. The best three activation functions found show a very large improvement of around 6% in precision on the magic telescope dataset, while also obtaining a higher accuracy and somewhat similar recall as shown in Table 4. Table 5 shows that the best functions found outperform the baseline functions on

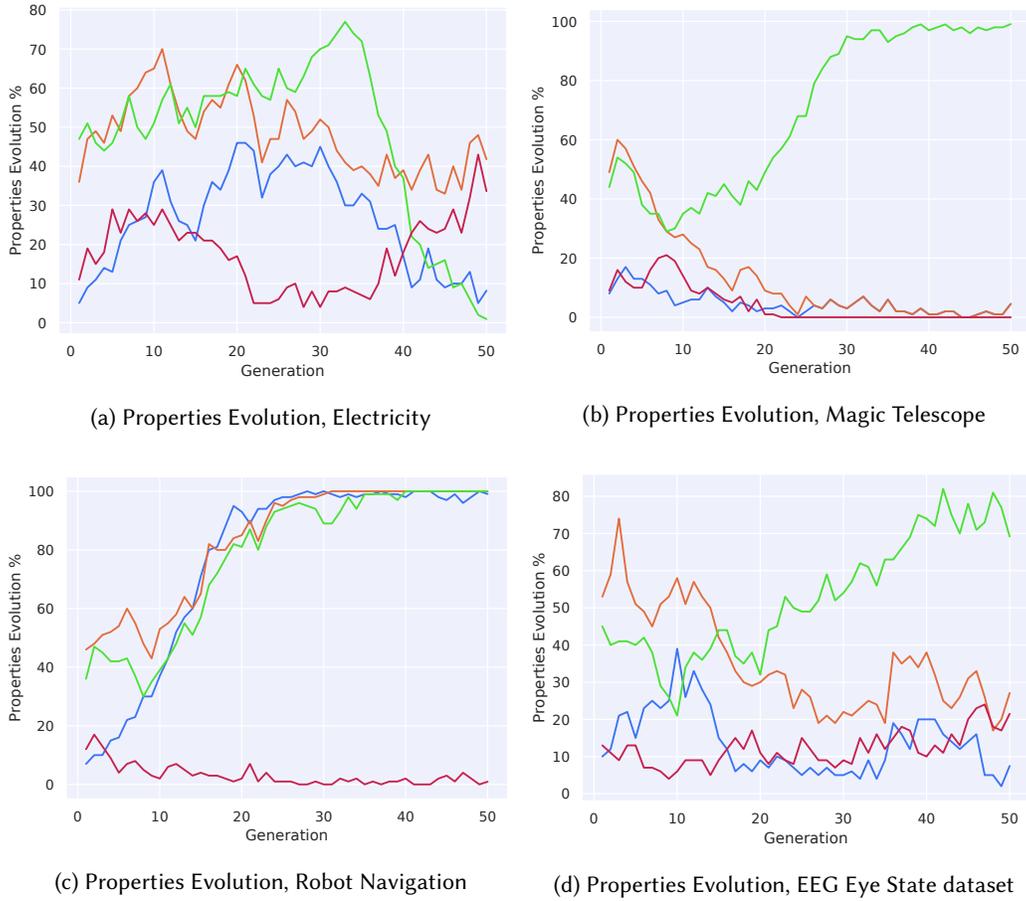

(a) Properties Evolution, Electricity

(b) Properties Evolution, Magic Telescope

(c) Properties Evolution, Robot Navigation

(d) Properties Evolution, EEG Eye State dataset

Fig. 7. The color denotes the property that is being graphed:

— — — : Zero on nonnegative real axis

— — — : Monotonically nondecreasing

— — — : Upper unbounded

— — — : Lower unbounded

Table 3. Results, Electricity Dataset

| Activation Function | Weight Initialization | Precision | Recall | $F_1$ Measure | Accuracy |
|---|---|---|---|---|---|
| $f_1$ | Glorot Uniform | **78.09% (1.67%)** | 87.92% (2.44%) | **82.67% (0.25%)** | **78.82% (0.45%)** |
| $f_2$ | He Normal | 76.9% (0.91%) | 88.76% (1.5%) | 82.39% (0.21%) | 78.19% (0.29%) |
| $f_3$ | Glorot Uniform | 77.33% (1.16%) | 88.2% (1.77%) | 82.38% (0.2%) | 78.32% (0.35%) |
| ReLU (baseline) | Glorot Uniform | 76.85% (1.27%) | **89.08% (1.88%)** | 82.48% (0.17%) | 78.26% (0.39%) |
| ELU (baseline) | He Normal | 75.6% (0.89%) | 86.1% (1.5%) | 80.5% (0.2%) | 76% (0.2%) |
| SELU (baseline) | LeCun Normal | 75.3% (0.4%) | 87.6% (1.0%) | 81% (0.2%) | 76.4% (0.1%) |

Table 4. Results, Magic Telescope Dataset

| Activation Function | Weight Initialization | Precision | Recall | $F_1$ Measure | Accuracy |
|---|---|---|---|---|---|
| $f_1$ | Glorot Normal | **78.65% (1.84%)** | 75.05% (1.98%) | 76.76% (0.44%) | **84.11% (0.3%)** |
| $f_2$ | He Normal | 77.91% (2.16%) | 75.56% (2.26%) | 76.66% (0.38%) | 83.91% (0.46%) |
| $f_3$ | Truncated Normal | 78.2% (2.07%) | 75.6% (2%) | **76.83% (0.42%)** | 84.05% (0.48%) |
| ReLU (baseline) | Glorot Uniform | 72% (1.86%) | **75.88% (1.41%)** | 73.86% (0.49%) | 81.21% (0.67%) |
| ELU (baseline) | He Normal | 66.8% (1.7%) | 74.8% (2.5%) | 70.5% (0.4%) | 78.1% (0.5%) |
| SELU (baseline) | LeCun Normal | 72.5% (2.4%) | 75.1% (1.9%) | 73.7% (0.6%) | 81.3% (0.8%)% |

Table 5. Results, EEG Eye State Dataset

| Activation Function | Weight Initialization | Precision | Recall | $F_1$ Measure | Accuracy |
|---|---|---|---|---|---|
| $f_1$ | Random Normal | 89.07% (1.34%) | 90.08% (1.52%) | 89.55% (0.54%) | 90.12% (0.51%) |
| $f_2$ | Random Normal | **89.37% (1.43%)** | **90.35% (1.67%)** | **89.84% (0.57%)** | **90.39% (0.54%)** |
| $f_3$ | Random Normal | 88.3% (1.46%) | 89.97% (1.82%) | 89.1% (0.57%) | 89.66% (0.52%) |
| ReLU (baseline) | Glorot Uniform | 86.85% (1.15%) | 88.54% (1.72%) | 87.67% (0.72%) | 88.3% (0.63%) |
| ELU (baseline) | He Normal | 85.3% (1.4%) | 87.8% (1.3%) | 86.5% (0.6%) | 87.1 % (0.6%) |
| SELU (baseline) | LeCun Normal | 84.8 % (1.4%) | 86.3% (2%) | 85.5% (0.7%) | 86.3% (0.6%) |

Table 6. Results, Robot Navigation Dataset

| Activation Function | Weight Initialization | Accuracy |
|---|---|---|
| $f_1$ | Random Normal | **92.07% (0.66%)** |
| ReLU (baseline) | Glorot Uniform | 91.21% (0.59%) |
| ELU (baseline) | He Normal | 90.3% (0.7%) |
| SELU (baseline) | LeCun Normal | 90.7% (0.7%) |

every metric on the EEG Eye State dataset. Finally, it is interesting to note that while the algorithm converged to a single function, the ReLU, for the Robot Navigation dataset, it succeeded in finding a weight initialization scheme that is not commonly associated with ReLU, but which outperformed the glorot uniform scheme on this particular dataset, as shown in Table 6.

*4.1.2 Statistical Analysis:* We analyze the significance of the difference in the performance of the activation functions by conducting the Tukey Honestly Significant Difference (HSD) statistical test with a confidence level of 95% Results reported in Table 7 show that, on the Electricity data set, $f_1$ significantly outperforms all the baseline functions on precision, $F_1$ measure and accuracy with a $p - value$ < 0.001 in all cases except when compared to ReLU on the $F_1$ measure where the $p - value$ equals 0.005 which still indicates statistical significance in the improvement. Moreover, the improvement that ReLU shows in recall over all the evolved functions is not statistically significant ($p - value$ > 0.005). On the Magic Telescope data set, $f_1$ significanlty outperforms all the baseline functions on precision and accuracy while $f_3$ outperforms them significantly on $F_1$ measure. Moreover, the improvement of ReLU over the evolved functions on recall is not significant ($p - value$ > 0.005). On EEG Eye data set, the improvement of $f_2$ over the baseline functions is statistically significant on all metrics with a $p - value$ <0.001. On the Robot Navigation data, $f_1$ significantly outperforms all base line functions with a $p - value$ <0.001.

*4.1.3 Analysis of the Evolution of Activation Function Properties on the Multi-variate Classification Datasets:* We now study the evolution of the activation properties for the datasets. The evolutions are shown in figures 7a, 7b, 7c, and 7d. As an example of how the study of the evolution of activation

Table 7. $p-values$ obtained using the Tukey HSD statistical test with 95% confidence comparing functions pairwise on the multi-variate data sets. $f_1$, $f_2$ and $f_3$ are different across the data sets.

| | Electricity | | | | Magic Telescope | | | | EEG Eye State | | | | Robot Navigation |
|---|---|---|---|---|---|---|---|---|---|---|---|---|---|
| Activation Function | Precision | Recall | F1 measure | Accuracy | Precision | Recall | F1 measure | Accuracy | Precision | Recall | F1 measure | Accuracy | Accuracy |
| $f_1$ vs ReLU | < .001 | .1 | .006 | < .001 | .613 | < .001 | < .001 | < .001 | .006 | < .001 | < .001 | < .001 | < .001 |
| $f_1$ vs ELU | < .001 | .001 | < .001 | < .001 | .996 | < .001 | < .001 | < .001 | < .001 | < .001 | < .001 | < .001 | < .001 |
| $f_1$ vs SELU | < .001 | .98 | < .001 | < .001 | 1 | < .001 | < .001 | < .001 | < .001 | < .001 | < .001 | < .001 | < .001 |
| $f_2$ vs ReLU | .999 | .98 | .54 | .957 | .99 | < .001 | < .001 | < .001 | < .001 | < .001 | < .001 | < .001 | |
| $f_2$ vs ELU | < .001 | < .001 | < .001 | < .001 | .699 | < .001 | < .001 | < .001 | < .001 | < .001 | < .001 | < .001 | |
| $f_2$ vs SELU | < .001 | .106 | < .001 | < .001 | .952 | < .001 | < .001 | < .001 | < .001 | < .001 | < .001 | < .001 | |
| $f_3$ vs ReLU | .56 | .369 | .419 | .978 | .994 | < .001 | < .001 | < .001 | .015 | < .001 | < .001 | < .001 | |
| $f_3$ vs ELU | < .001 | < .001 | < .001 | < .001 | .65 | < .001 | < .001 | < .001 | < .001 | < .001 | < .001 | < .001 | |
| $f_3$ vs SELU | < .001 | .763 | < .001 | < .001 | .932 | < .001 | < .001 | < .001 | < .001 | < .001 | < .001 | < .001 | |

properties can be useful, consider the plot in Figure 7c. We can see that less than 10% of the individuals at generation 1 had the zero on non-negative axis property, and that percentage steadily increased to 100% of the individuals having that property by the last generation. This provides good evidence that this property is useful for this particular dataset and architecture (although we have to take into account the caveats discussed in Section 3.3). In general, Figure 7c shows that on the Robot Navigation dataset, the evolutionary algorithm favors activation functions that have the zero on non-negative axis property, that are upper unbounded and monotonically non-decreasing but are not lower unbounded. That is, the evolutionary algorithm favors activation functions that look exactly like the ReLU. Figure 7a shows that the evolutionary algorithm appears to favor activation functions that are not upper unbounded on the Electricity dataset, which is reflected in the fact that the functions in Figure 5a are all bounded on $\mathbb{R}^+$. The lower unboundedness property achieves a fair degree of prominence in the activation functions found on the Electricity dataset, which around 35% of the activation functions being lower unbounded by the final generation. Monotonicity also appears to be somewhat useful, while the zero on non-negative real axis property achieved some prominence in in the 20th to 30th generation but was filtered out by the end. Figures 7b, and 7d indicate that the algorithm shows some similar trends between the datasets, such as the fact that it favors activation functions that are upper unbounded but that do not have the zero on non-negative axis property or the lower unboundedness property. However, the evolutionary algorithm holds on to some individuals that monotonically non-decreasing in the EEG Eye State experiments (around 20-30%), while it filters them out completely in the Magic Telescope experiment. It is interesting to see that while monotonic activation functions are generally considered to be useful, the favoritism for them is not as strong as one would expect in our experiments. In addition, the potential sparsity offered by a hard zero does not appear to be a generally useful property.

### 4.2 Results on Regression Datasets

We choose three datasets for the regression experiments, which are the Red Wine Quality, White Wine Quality, and California Housing datasets. The Red Wine Quality and White Wine Quality datasets can be obtained from the UCI Machine Learning repository, while the California Housing dataset comes packaged with the scikit-learn python package. We used a 75/25 train/test split for the three datasets. A description of the datasets is given in Table 8, and the architectures of the neural networks used for each dataset are described in Table 9.

*4.2.1 Best Three Functions Found for the Regression Datasets:* The plots of the best activation functions found on the regression experiments are shown in figures 8a, 8b, and 8c. We can see that the activation functions found on the regression datasets do not look anything like the ones found on the multi-variate classification experiments. All of the functions in figures 8a, 8b, and 8c are bounded, and the function values fall in the interval [-1,4], which are much smaller values than the ones which the functions in figures 5a, 5b, 5c and 5d take on. There seems to be a trend of incorporating sines and cosines in the activation functions, which is an extremely unexpected result because the periodic and non-monotonic nature of the functions would be expected to "confuse" the back-propagation process, in the sense that decreasing the weight for a neuron might actually increase its importance and vice versa for increasing the weight. $f_2$ in 8a is also particularly surprising as it changes direction extremely quickly, and it looks to be a particularly bad fit for back-propagation, to the point that one would expect it to result in a random search for weights. However, the empirical results indicate otherwise for this dataset. The results for the regression experiments are shown in tables 10, 11, and 12 and Figure 9 . We observe that the evolutionary algorithm succeeded in finding activation functions that achieve a lower mean squared error than the baseline functions on all the regression datasets.

Table 8. Description of the datasets used for the regression experiments

| Dataset | № Instances | № Features |
|---|---|---|
| Red Wine Quality | 1599 | 12 |
| White Wine Quality | 4898 | 12 |
| California Housing | 20640 | 8 |

Table 9. Experimental parameters for the regression datasets

| Dataset | № Hidden Layers | № Neurons per Layer | Dropout Rate | $L_2$ Regularization | Scaling Method |
|---|---|---|---|---|---|
| Red Wine Quality | 1 | 40 | 30% | Yes | MinMaxScaling |
| White Wine Quality | 2 | 50 | 0% | No | StandardScaling |
| California Housing | 5 | 50 | 40% | No | MinMaxScaling |

## Results on the Regression Datasets

Table 10. Results, Red Wine Quality

| Activation Function | Weight Initialization | Mean Squared Error |
|---|---|---|
| $f_1$ | He Uniform | 0.506 (0.02) |
| $f_2$ | LeCun Uniform | 0.496 (0.017) |
| $f_3$ | LeCun Uniform | **0.485 (0.01)** |
| ReLU (baseline) | Glorot Uniform | 0.818 (0.032) |
| ELU (baseline) | He Normal | 0.761 (0.044) |
| SELU (baseline) | LeCun Normal | 0.746 (0.044) |

Table 11. Results, White Wine Quality

| Activation Function | Weight Initialization | Mean Squared Error |
|---|---|---|
| $f_1$ | He Uniform | 0.584 (0.013) |
| $f_2$ | Truncated Normal | **0.578 (0.012)** |
| $f_3$ | He Uniform | 0.59 (0.01) |
| ReLU (baseline) | Glorot Uniform | 0.753 (0.032) |
| ELU (baseline) | He Normal | 0.779 (0.04) |
| SELU (baseline) | LeCun Normal | 0.787 (0.041) |

Table 12. Results, California Housing

| Activation Function | Weight Initialization | Mean Squared Error |
|---|---|---|
| $f_1$ | Glorot Normal | **0.399 (0.006)** |
| $f_2$ | Glorot Normal | 0.417 (0.003) |
| $f_3$ | Glorot Normal | 0.417 (0.005) |
| ReLU (baseline) | Glorot Uniform | 0.444 (0.019) |
| ELU (baseline) | He Normal | 0.431 (0.006) |
| SELU (baseline) | LeCun Normal | 0.416 (0.008) |

*4.2.2 Statistical Analysis:* Results reported in Table 13 show the p-values obtained with the Tukey test conducted on the regression data sets with a confidence level of 95%. Results on the Red Wine Quality data set show that $f_3$ significantly statistically outperforms the base line functions with Tukey HSD statistical test leading a $p-value$ <0.001. On the White Wine Quality data, the

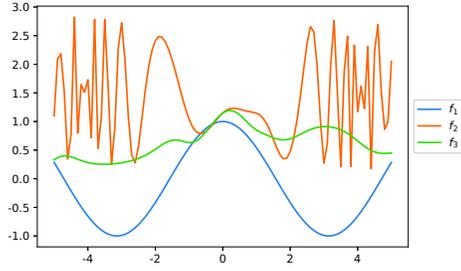

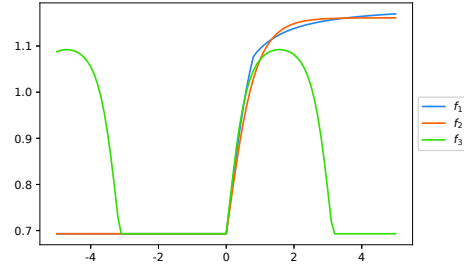

(a) Plot of the best 3 activation functions found on the Red Wine Quality dataset

- $f_1(x) = \cos(x)$

- $f_2(x) = \cosh(\cos(\cos(\cos(\text{atan}(x) \cdot \cos(\cos^2(x) + \cos(x))))) + \text{atan}(x) \cdot \cos(\cosh(x) + \text{atan}(\cos(\text{elu}(\text{elu}(x))) \cdot \cos^2(x) \cdot \text{atan}(x))$

- $f_3(x) = \cosh(\cos(\cos(\text{atan}(x) \cdot \cos(\cos^2(x) + \cos(x))))) + \text{atan}(x) \cdot \cos(\text{atan}(\text{atan}(x)) + \cosh(\cos(\cos(x))))$

(b) Plot of the best 3 activation functions found on the White Wine Quality dataset

- $f_1(x) = \text{softplus}(\min(\text{erf}(\text{relu}(\text{atan}(x))), \text{sigmoid}(\text{relu}(\text{atan}(x)))))$

- $f_2(x) = \text{softplus}(\text{relu}(\text{atan}(\text{relu}(\text{tanh}(x)))))$

- $f_3(x) = \text{softplus}(\text{erf}(\text{relu}(\text{erf}(\text{erf}(\sin(\sin(\sin(x)))))))))$

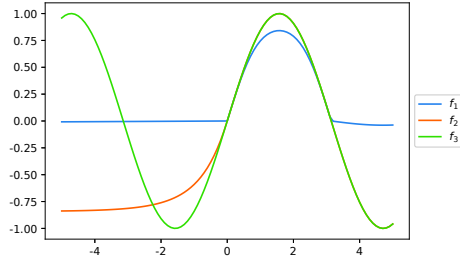

(c) Plot of the best 3 activation functions found on the California Housing dataset

- $f_1(x) = \sin(\text{leaky-relu}(\text{leaky-relu}(\sin(\text{leaky-relu}(x)))))$

- $f_2(x) = \sin(\text{elu}(x))$

- $f_3(x) = \sin(x)$

Fig. 8. Plots of the best activation functions found for the regression problems.

improvement that $f_2$ shows over all all baseline functions is statistically significant with a p value <0.001. On the California Housing data set, $f_1$ shows an improvement over all baseline functions that is statistically significant ($p - value < 0.001$).

### 4.2.3 Analysis of the Evolution of Activation Function Properties on the Regression Datasets:
The plots for the evolution of the activation function properties are shown in figures 10a, 10b, and 10c. The properties that are deemed to be useful by the evolutionary algorithm on the regression datasets are very different from the ones deemed useful by the algorithm on most of the classification datasets. For example, activation functions that are upper unbounded are quickly filtered out of the population on all three datasets. In the experiment on the Red Wine Quality dataset, almost 50% of the initial activation functions were upper unbounded, but the algorithm filtered them out completely by the end, as shown in Figure 10a. The zero on non-negative real axis property does not

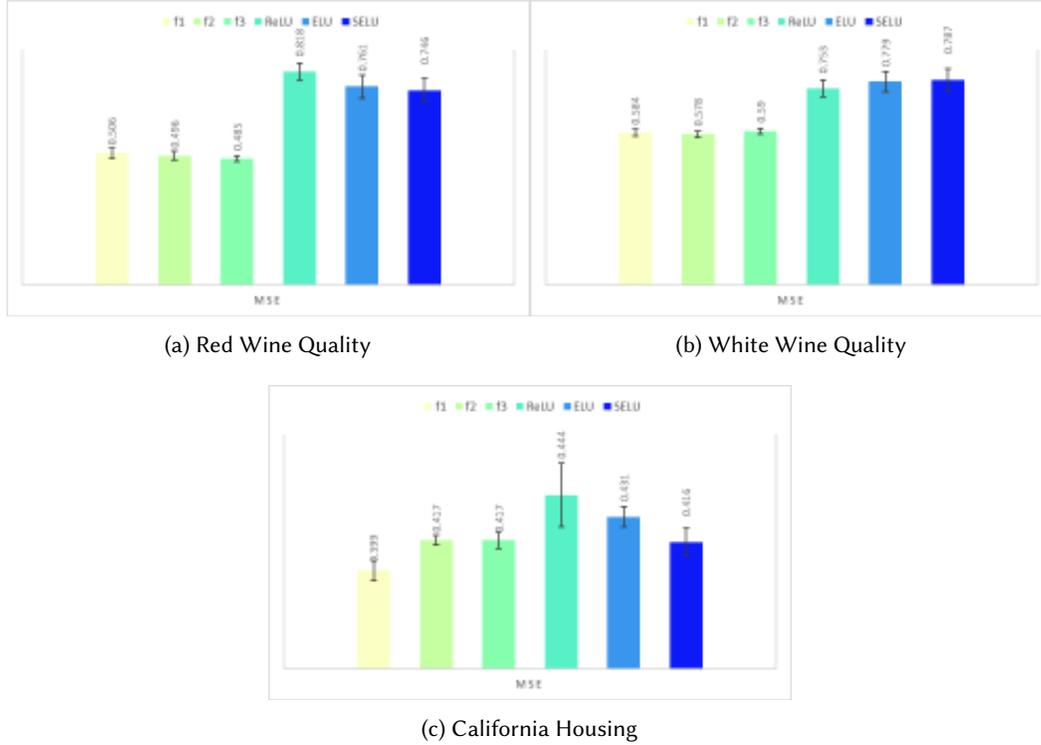

(a) Red Wine Quality

(b) White Wine Quality

(c) California Housing

Fig. 9. Bar plots showing results (Mean Squared Error) on the regression datasets

Table 13. p-values obtained using the Tukey HSD statistical test with 95% confidence comparing functions pairwise on the regression data sets. $f_1$, $f_2$ and $f_3$ are different across the data sets.

|  | Red Wine Quality | White Wine Quality | California Housing |
|---|---|---|---|
| **Activation Function** | **Mean Squared Error** | **Mean Squared Error** | **Mean Squared Error** |
| $f_1$ **vs ReLU** | < .001 | < .001 | < .001 |
| $f_1$ **vs ELU** | < .001 | < .001 | < .001 |
| $f_1$ **vs SELU** | < .001 | < .001 | < .001 |
| $f_2$ **vs ReLU** | < .001 | < .001 | < .001 |
| $f_2$ **vs ELU** | < .001 | < .001 | < .001 |
| $f_2$ **vs SELU** | < .001 | < .001 | .998 |
| $f_3$ **vs ReLU** | < .001 | < .001 | < .001 |
| $f_3$ **vs ELU** | < .001 | < .001 | < .001 |
| $f_3$ **vs SELU** | < .001 | < .001 | .998 |

appear to be useful: it was present in around 10% of the initially randomly generated individuals on the Red Wine Quality, White Wine Quality, and California Housing datasets, but the evolutionary algorithm filtered it out quickly. It has to be noted that $f_1$ on the California Housing dataset looks like it might have this property, but in fact, it takes on values which are extremely close to zero but not exactly zero on $\mathbb{R}^-$. In fact, for the Red Wine Quality and California Housing datasets, all of the activation function properties being studied were filtered out by the algorithm, as none of

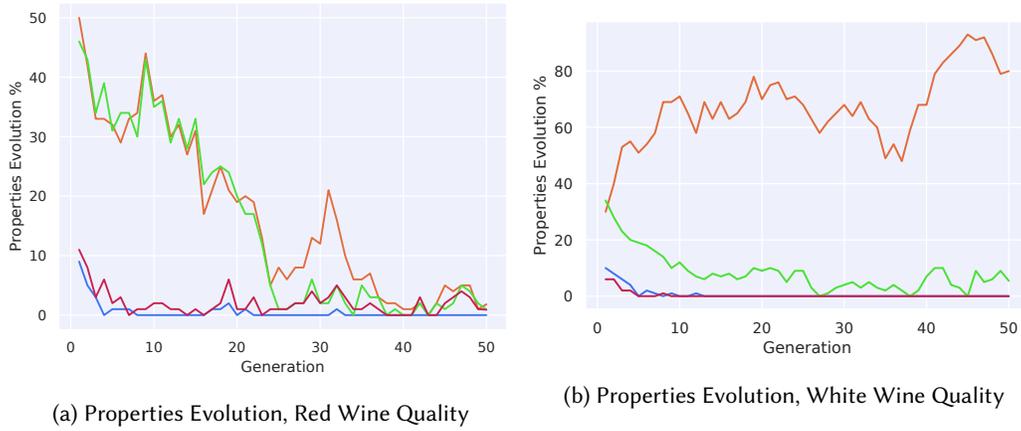

(a) Properties Evolution, Red Wine Quality

(b) Properties Evolution, White Wine Quality

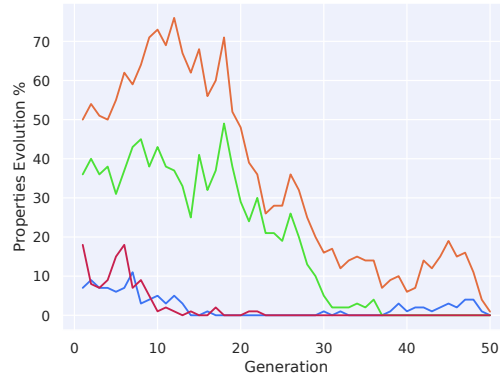

(c) Properties Evolution, California Housing

Fig. 10. The color denotes the property that is being graphed:

- — — — : Zero on nonnegative real axis
- — — — : Monotonically nondecreasing
- — — — : Upper unbounded
- — — — : Lower unbounded

them captures the unusual shape of the functions which perform well on this dataset, such as the functions shown in Figure 8a which incorporate a cosine. This is not the case for the White Wine Quality dataset, where one of the activation function properties being studied turned out to be useful, namely that of the monotonically non-decreasing property as shown in Figure 10b.

## 4.3 Results on Image-Based Datasets

We test our algorithm on 3 well known image-based datasets: the CIFAR-10 dataset [Krizhevsky et al. 2009], the Fashion-MNIST dataset [Xiao et al. 2017], and the MNIST dataset [LeCun et al. 2010]. Unfortunately, image-based datasets require the use of convolutional neural networks, which are much more computationally expensive to train. We do not have access to the computational power needed to run a full experiment on these datasets, so we only use a subset of 5000 examples for each of them. For reproducibility purposes, the subsets are obtained by first concatenating the train and test sets into a single dataset $D$, shuffling $D$ using the scikit-learn shuffle function with a random seed of 42, then splitting $D$ into 75/25 training/test sets using the scikit-learn train_test_split function with a random seed of 42. The only preprocessing done is encoding the target classes using a one-hot encoding scheme and dividing the pixel values by 255 to scale the values to the [0,1] range and the loss function chosen for all of the datasets is the categorical cross-entropy. We use a "same" padding for all convolutional layers and a "valid" padding for all pooling layers. The architectures used are summarized in tables 14, 15 and 16.

Architectures used on the image-based datasets

Table 14. CIFAR-10 architecture

| Layer | Layer Type | Number of filters or units | Filter size |
|---|---|---|---|
| 1 | Convolution | 16 | $3 \times 3$ |
| 2 | Max Pooling | N/A | $2 \times 2$ |
| 3 | Convolution | 32 | $3 \times 3$ |
| 4 | Max Pooling | N/A | $2 \times 2$ |
| 5 | Convolution | 64 | $3 \times 3$ |
| 6 | Max Pooling | N/A | $2 \times 2$ |
| 7 | Fully Connected | 128 | N/A |
| 8 | Fully Connected | 10 | N/A |

Table 15. Fashion-MNIST architecture

| Layer | Layer Type | Number of filters or units | Filter size |
|---|---|---|---|
| 1 | Convolution | 16 | $5 \times 5$ |
| 2 | Max Pooling | N/A | $2 \times 2$ |
| 3 | Dropout (25%) | N/A | N/A |
| 4 | Convolution | 32 | $5 \times 5$ |
| 5 | Max Pooling | N/A | $2 \times 2$ |
| 6 | Dropout (25%) | N/A | N/A |
| 7 | Fully Connected | 512 | N/A |
| 8 | Fully Connected | 10 | N/A |

Table 16. MNIST architecture

| Layer | Layer Type | Number of filters or units | Filter size |
|---|---|---|---|
| 1 | Convolution | 6 | $3 \times 3$ |
| 2 | Average Pooling | N/A | $2 \times 2$ |
| 3 | Convolution | 16 | $3 \times 3$ |
| 4 | Average Pooling | N/A | $2 \times 2$ |
| 5 | Fully Connected | 120 | N/A |
| 6 | Fully Connected | 84 | N/A |
| 7 | Fully Connected | 10 | N/A |

*4.3.1 Best Three Functions Found for the Image Classification Datasets:* The plots of the best activation functions found for the image-based classification datasets are shown in figures 11a, 11b, 11c. First of all, it is interesting to note that most of the functions in figures 11a and 11b have a value of zero on $\mathbb{R}^-$, just like the ReLU. The work in [Bingham et al. 2020; Nader and Azar 2020] shows that the evolutionary algorithm converged to functions that do not have this property when the full CIFAR-10 dataset is used, so it is possible that this property is only useful when little data is available. The three functions illustrated in Figure 11a look fairly similar to each other when $x \geq 0$, so it is clear that the algorithm is favoring a very specific type of function on this interval. As shown in Figure 11b, the algorithm converged to two activation functions on the Fashion MNIST dataset, one of which looks like the activation functions found for the CIFAR-10 dataset, while the other is the baseline ReLU function. Again, this provides good evidence that our algorithm is able to recover baseline functions if they perform well for a particular dataset. The functions found in Figure 11c are probably the most interesting in this entire study. The three functions are almost exactly equal in a small interval $I$ around $x = 0$, but they look nothing alike outside of this interval. When $x \notin I$, $f_1$ is zero while $f_3$ takes on strictly positive values and $f_2$ takes on a periodic look. A reasonable explanation for this phenomenon is that most of the weights of the neural network are small, which means that the input to the activation functions of each neuron does not fall outside of the interval $I$, hence explaining why the shape of the activation functions does not look to be important outside of this interval. Looking at Table 17, we can see that the algorithm succeeds

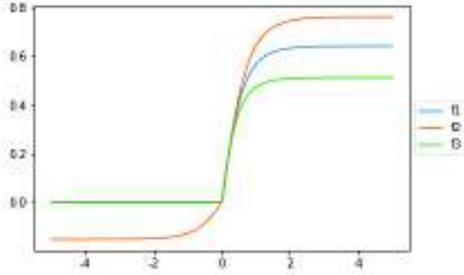

(a) Plot of the best 3 activation functions found on the CIFAR-10 subset

- $f_1(x) = \text{leaky-relu}(\tanh(\text{relu}(\tanh(\text{relu}(\\ \tanh(\text{relu}(x)))))))$
- $f_2(x) = \text{leaky-relu}(\tanh(\tanh(x)))$
- $f_3(x) = \tanh(\text{relu}(\tanh(\tanh(\tanh(\tanh(\tanh(x)))))))$

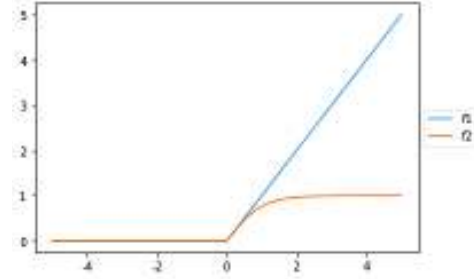

(b) Plot of the best 2 activation functions found on the Fashion MNIST subset

- $f_1(x) = \text{relu}(x)$
- $f_2(x) = \text{relu}(\tanh(x))$

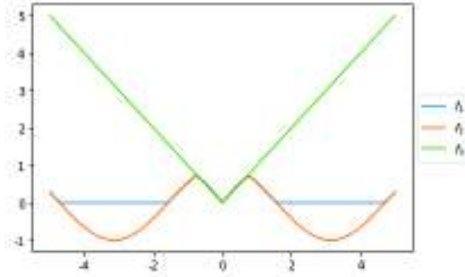

(c) Plot of the best 3 activation functions found on the MNIST subset

- $f_1(x) = \min(\text{relu}(\cos(x)), |\text{erf}(x)|)$
- $f_2(x) = \min(\cos(x), |\text{erf}(|x|)))$
- $f_3(x) = |x|$

Fig. 11. Plots of the best activation functions found for the image-based classification problems.

in finding activation functions that obtain an improvement of around 3% in accuracy over the baseline functions on the CIFAR-10 dataset, although the accuracy of all functions is not very high because the convolutional neural network was only trained on 5000 examples. As Table 18 shows, one of the best functions found by the algorithm on the Fashion MNIST dataset was the ReLU function although its associated weight initialization scheme resulted in slightly lower accuracy than the baseline, and the second function found by the algorithm on the Fashion MNIST dataset is the $\text{relu}(\tanh(x))$, which performs better than the baselines, although again, the difference is not too drastic. Finally, Table 19 shows that the three activation functions found perform almost similarly and achieve an improvement of around 0.7% over the best baseline. Figure 12 summarizes the results.

*4.3.2 Statistical Analysis:* Table 20 shows the p-values obtained with the Tukey statistical test conducted on the image-based data sets. A $p-value < 0.001$ on the CIFAR-10 data indicates that the improvement of $f_3$ over the baseline functions is statistically significant. A similar result indicates

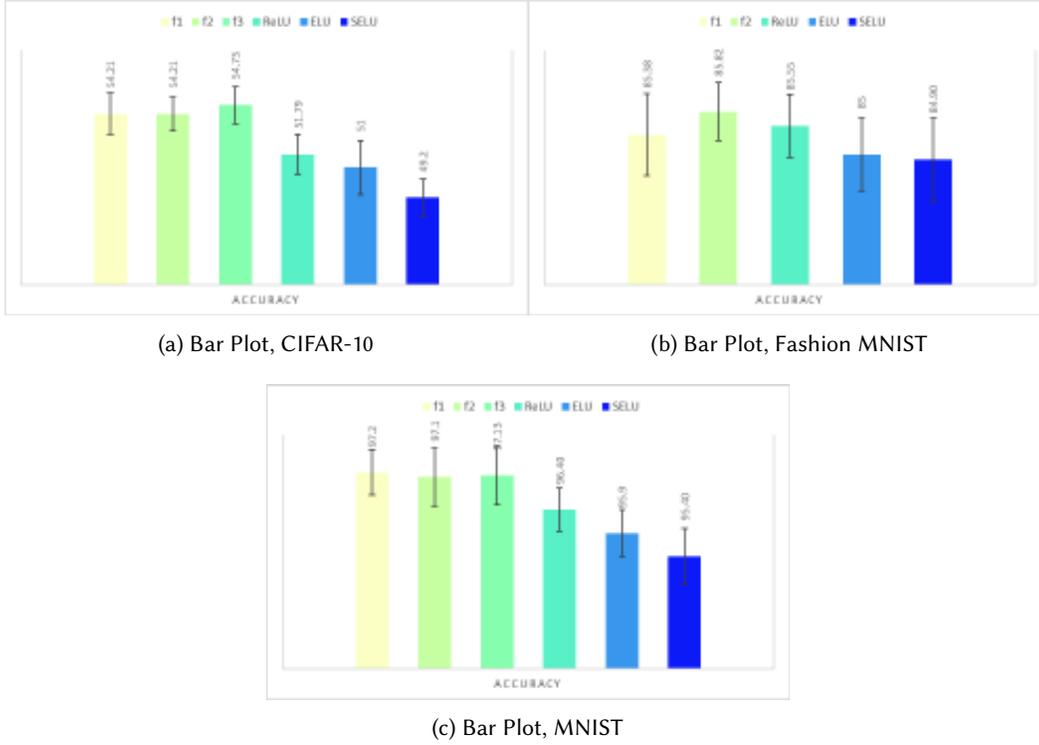

(a) Bar Plot, CIFAR-10          (b) Bar Plot, Fashion MNIST

(c) Bar Plot, MNIST

Fig. 12.  Bar plots showing results (Accuracy) on the image-based datasets

Results on the image-based Classification Datasets

Table 17.  Results, CIFAR-10 subset

| Activation Function | Weight Initialization | Accuracy |
|---|---|---|
| $f_1$ | LeCun Uniform | 54.21 % (1.23%) |
| $f_2$ | LeCun Uniform | 54.21% (1%) |
| $f_3$ | Orthogonal | **54.73% (1.1%)** |
| ReLU (baseline) | Glorot Uniform | 51.79% (1.18%) |
| ELU (baseline) | He Normal | 51% (1.6%) |
| SELU (baseline) | LeCun Normal | 49.2% (1.1%) |

Table 18.  Results, Fashion MNIST subset

| Activation Function | Weight Initialization | Accuracy |
|---|---|---|
| ReLU | Random Uniform | 85.38%(0.78%) |
| $f_2$ | LeCun Normal | **85.82%(0.56%)** |
| ReLU (baseline) | Glorot Uniform | 85.55%(0.6%) |
| ELU (baseline) | He Normal | 85% (0.7%) |
| SELU (baseline) | LeCun Normal | 84.9% (0.8%) |

Table 19.  Results, MNIST subset

| Activation Function | Weight Initialization | Accuracy |
|---|---|---|
| $f_1$ | Glorot Uniform | **97.2% (0.49%)** |
| $f_2$ | Orthogonal | 97.1% (0.62%) |
| $f_3$ | Orthogonal | 97.13 % (0.62%) |
| ReLU (baseline) | Glorot Uniform | 96.4%(0.46%) |
| ELU (baseline) | He Normal | 95.9% (0.5%) |
| SELU (baseline) | LeCun Normal | 95.4% (0.6%) |

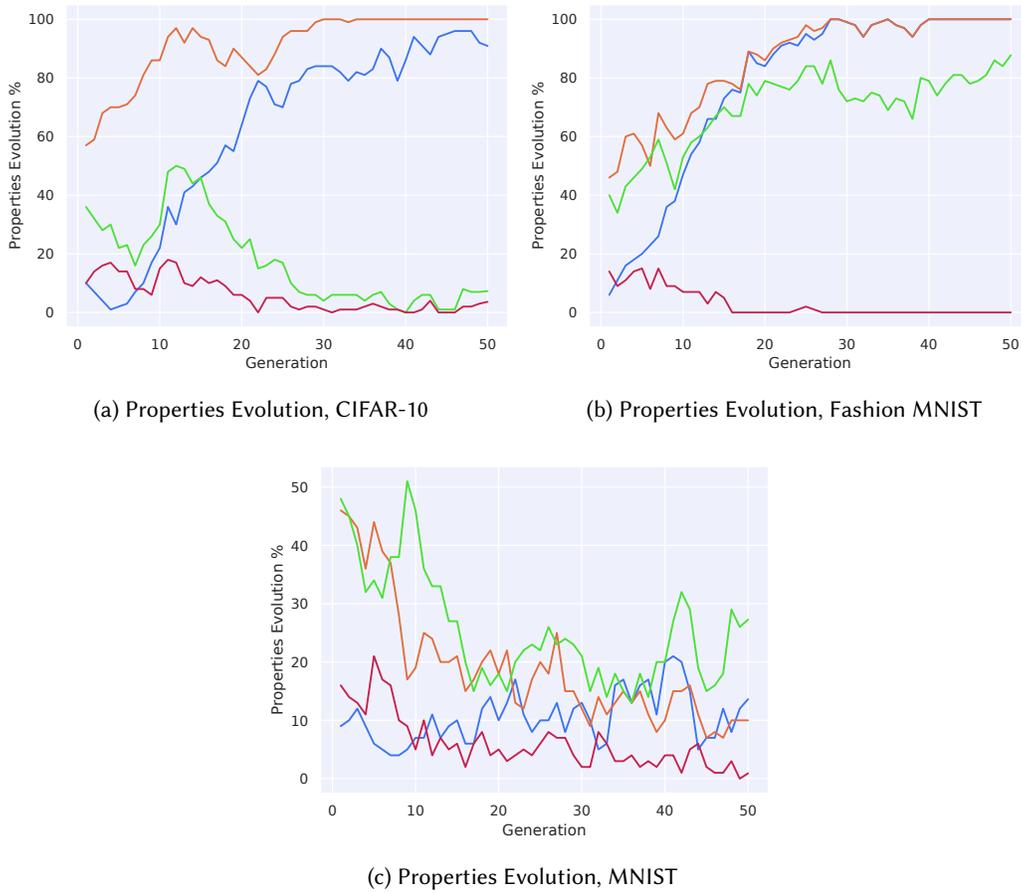

(a) Properties Evolution, CIFAR-10

(b) Properties Evolution, Fashion MNIST

(c) Properties Evolution, MNIST

Fig. 13. The color denotes the property that is being graphed:

----- : Zero on nonnegative real axis

----- : Monotonically nondecreasing

----- : Upper unbounded

----- : Lower unbounded

that $f_1$ statistically significantly outperforms the base line functions on MNIST. On the Fashion MNIST data, the improvement of $f_2$ over ELU and SELU is statistically significant but not the one over ReLU(pvalue =0.56).

*4.3.3 Analysis of the Evolution of Activation Function Properties for the Image-Based Classification Datasets:* We now study the evolution of the activation function properties for the image-based datasets. The evolutions are shown in figures 13a, 13b, and 13c. We can see that the algorithm is clearly favoring activation functions that are monotonically non-decreasing and that have a zero on the non-negative real axis for the CIFAR-10 and Fashion-MNIST datasets, as shown in figures 13a and 13b. The algorithm only exhibited this type of favoritism for the zero on non-negative real axis property once before, on the Robot Navigation dataset as shown in Figure 7c. Even though the properties being favored by the algorithm on the CIFAR-10 and Fashion-MNIST datasets share some commonalities, there are also some differences, notably the fact that the upper unbounded property gradually disappears from the population on the CIFAR-10 dataset while it is heavily favored on the Fashion-MNIST dataset. Finally, Figure 13c shows that almost all of the properties completely disappeared from the population by the final generation, with the exception of the upper unbounded property, which stays in around 30% of the individuals in the final generation. These properties disappeared because none of them capture the very unusual shape of the activation functions which perform well on this dataset, such as the ones shown in Figure 11c.

Table 20. p-values obtained using the Tukey HSD statistical test with 95% confidence comparing functions pairwise on the image-based data sets. $f_1$, $f_2$ and $f_3$ are different across the data sets.

| | CIFAR-10 | MNIST | Fashion MNIST |
|---|---|---|---|
| **Activation Function** | **Accuracy** | **Accuracy** | **Accuracy** |
| $f_1$ **vs ReLU** | $< .001$ | $< .001$ | .877 |
| $f_1$ **vs ELU** | $< .001$ | $< .001$ | .217 |
| $f_1$ **vs SELU** | $< .001$ | $< .001$ | .062 |
| $f_2$ **vs ReLU** | $< .001$ | $< .001$ | .56 |
| $f_2$ **vs ELU** | $< .001$ | $< .001$ | $< .001$ |
| $f_2$ **vs SELU** | $< .001$ | $< .001$ | $< .001$ |
| $f_3$ **vs ReLU** | $< .001$ | $< .001$ | |
| $f_3$ **vs ELU** | $< .001$ | $< .001$ | |
| $f_3$ **vs SELU** | $< .001$ | $< .001$ | |

## 5 CONCLUSIONS AND FUTURE WORK

We have shown that it is feasible to use an evolutionary algorithm based on a genetic programming approach to search for activation functions. The results were positive, with the algorithm finding activation functions that outperform statistically significantly the baseline functions on some datasets, such as the Electricity and Magic Telescope datasets (as shown in Tables 3 and 4). It is also promising to see that if the baseline functions themselves are close to optimal, then the algorithm is capable of recovering them, as shown in figures 5c and 11b where the algorithm chose the ReLU function. These observations mean that the algorithm is extremely well suited to being integrated as part of a Neural Architecture Search algorithm. Neural Architecture Search algorithms usually focus on evolving new architectures, but we have shown that it is worth adding the search for new activation functions to the mix. When it comes to the shape of the activation functions being evolved, we have studied the evolution of various properties such as monotonicity and found that the algorithm favored different kinds of functions for each experiment. This implies that the commonly held heuristics (monotonicity, zero on non-negative real axis, upper unboundedness, lower unboundedness) about what an activation function should look like do not hold. An interesting line of avenue for future work would be to run the NAS algorithm on the same dataset and with the same parameters multiple times, and see if the functions found by the algorithm differ drastically between each run. This would shed more light on whether the algorithm is favoring a certain type of function because of some early bias in its search, or if it is favoring this type because it is the best one for the problem at hand. Such a study would be beneficial for understanding the degree of diversity of the activation functions that perform well on a particular dataset. Unfortunately, a lack of access to computational power meant that we could not run our algorithm on large modern architectures, and we could not use the full image-based datasets. An obvious future line of work would thus be to do just that, and test the algorithm on extremely large architectures and datasets. Finally, we note that we have focused on activation functions that perform well without trying to favor functions that are not computationally expensive to optimize. In order to make the algorithm more practical, we think that it would be extremely useful to think of ways to favor simpler activation functions without imposing any particular restraints on their shape.